\RequirePackage{amsthm}

\documentclass[sn-mathphys,Numbered]{sn-jnl}

\usepackage{graphicx}%
\usepackage{multirow}%
\usepackage{amsmath,amssymb,amsfonts}%
\usepackage{mathrsfs}%
\usepackage[title]{appendix}%
\usepackage{xcolor}%
\usepackage{textcomp}%
\usepackage{manyfoot}%
\usepackage{booktabs}%
\usepackage{algorithm}%
\usepackage{algorithmicx}%
\usepackage{algpseudocode}%
\usepackage{listings}%

\theoremstyle{thmstyleone}%
\theoremstyle{thmstyletwo}%

\theoremstyle{thmstylethree}%

\begin{document}

\title[Article Title]{DeepArt: A Benchmark to Advance Fidelity Research in AI-Generated Content}


\author[1]{\fnm{Wentao} \sur{Wang}}\email{wentao.wang@ieee.org}
\equalcont{These authors contributed equally to this work.}

\author[1]{\fnm{Xuanyao} \sur{Huang}}\email{xuanyao.huang@ieee.org}
\equalcont{These authors contributed equally to this work.}

\author[2]{\fnm{Tianyang} \sur{Wang}}\email{tw2@uab.edu}

\author*[3]{\fnm{Swalpa Kumar} \sur{Roy}}\email{swalpa@cse.jgec.ac.in}

\affil[1]{\orgdiv{School of Automation}, \orgname{Chongqing University of Posts and Telecommunications}, \orgaddress{\city{Chongqing}, \postcode{400065}, \state{Chongqing}, \country{China}}}

\affil[2]{\orgdiv{Department of Computer Science}, \orgname{University of Alabama at
Birmingham}, \orgaddress{\city{Birmingham}, \postcode{35294}, \state{AL}, \country{USA}}}

\affil*[3]{\orgdiv{Department of Computer Science and Engineering}, \orgname{Alipurduar Government Engineering and Management College}, \orgaddress{\city{Alipurduar}, \postcode{736206}, \state{West Bengal}, \country{India}}}


\abstract{This paper explores the image synthesis capabilities of GPT-4, a leading multi-modal large language model. We establish a benchmark for evaluating the fidelity of texture features in images generated by GPT-4, comprising manually painted pictures and their AI-generated counterparts. The contributions of this study are threefold: First, we provide an in-depth analysis of the fidelity of image synthesis features based on GPT-4, marking the first such study on this state-of-the-art model. Second, the quantitative and qualitative experiments fully reveals the limitations of the GPT-4 model in image synthesis. Third, we have compiled a unique benchmark of manual drawings and corresponding GPT-4-generated images, introducing a new task to advance fidelity research in AI-generated content (AIGC). The dataset is available at: \url{https://github.com/rickwang28574/DeepArt}.}




\maketitle

\begin{figure}[h]%
\centering
\includegraphics[width=1\textwidth]{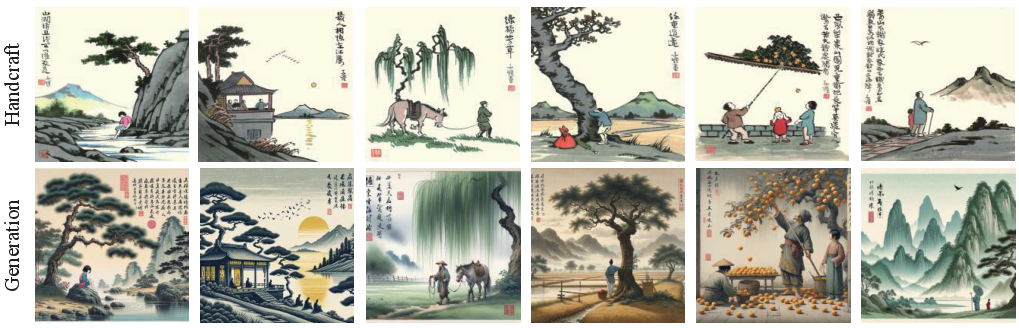}
\caption{Here, we show a range of prompt-guided generation images with an original handcraft. The presence of such images raises the research issues on artificial intelligence image synthesis fidelity. Top row: original handcraft. Bottom row: prompt-guided generation. Handcraft and generation are similar, and sometimes they are different.}\label{fig1}
\end{figure}

\section{Introduction}\label{sec1}

Multi-modal large language models (LLMs) represented by GPT-4 have shown powerful functions in generating corresponding synthetic images through text prompts~\cite{cao2023comprehensive,zhang2023complete,wu2023ai}. They can interpret the text descriptions provided by users and convert them for visual representation. Its power and ease of operation make it widely used in many fields, including but not limited to painting creation, product design~\cite{hong2023generative}, human-computer interaction~\cite{hamalainen2023evaluating}, medical analysis~\cite{waisberg2023gpt,nori2023capabilities}, etc. However, its powerful generation ability still has many flaws, and it is easy to lose, distort, and fabricate texture features to varying degrees.  While several studies have investigated the differences between images generated by artificial intelligence and those that are either natural or artificially created, most have focused primarily on specific image generation models like Stable Diffusion. Unlike specific generative networks, GPT-4 has broader applications in daily life in modern society, making it imperative to conduct more in-depth research on their image generation ability to align with societal needs and trends. However, existing work currently focuses on exploring the breadth of GPT-4's generation capabilities. That is, exploring how it performs in various application scenarios~\cite{mao2023gpteval,liu2023evaluating,sun2023evaluating,naismith2023automated}. Conversely, research on the fidelity of texture features in GPT-4 generated images is still less.

A very intuitive solution is to compare natural or manufactured images with images generated by GPT-4. However, the evaluation of this method must be conducted against specific benchmarks. It is worth noting that limited similar benchmarks are available for assessing the performance of large multi-modal models, such as GPT-4. Indeed, various large multi-modal models, exemplified by GPT-4, can generate synthetic images in bulk. Acquiring sufficient synthetic images is not a challenging task. However, it is essential to note that the image generation capability of these large models relies on extensive data pre-training, so the resulting model-generated images tend to exhibit a significant degree of randomness. This inherent randomness can make it difficult to establish a clear correspondence between the samples generated by the model and real-world images, which will cause certain uncertainties in the research.

To address this problem, this paper first introduces an original benchmark consisting of hand-drawn images. This batch of hand-painted images comes from the famous painter Feng Zikai. Represents genuine human artistic expression, possesses a high level of craftsmanship, is available through the Open Access Program, and may be legally and readily used for research purposes. Then, we proposed an "encoding-decoding" mapping method based on GPT-4, using this method to create synthetic data corresponding with the original data, thus forming pairs of corresponding and meaningful artificial data and generated images. At this point, we have successfully developed and defined a novel data benchmark: DeepArt. Finally, this paper conducts a preliminary quantitative and qualitative evaluation of the DeepArt benchmark, and proposes new challenges in the "image fidelity" of multi-modal large models represented by GPT-4.

The main contributions are four-fold: 1) To the best of our knowledge, we are the first to study the fidelity of features in image synthesis based on the current state-of-the-art large language model GPT-4. We use hand-drawn images as a starting point, deeply explore the features differences between images generated based on GPT-4 and artificial images. 2) We propose an "encoding-decoding" mapping method based on GPT-4, using this method to create synthetic data corresponding to the original data. This method is highly scalable and can be used as a reference for producing other types of data samples. 3) Quantitative and qualitative experiments reflect the shortcomings of the GPT4 large language model in image synthesis and demonstrate new challenges in the "image fidelity" for our defined benchmarks and missions. 4) We collected and constructed a dataset of artificial images and corresponding GPT-4 generated images. Provide open access at: \url{https://github.com/rickwang28574/DeepArt}.

\section{Related Works}

This section will first describe the relevant works produced by specific generative networks. Then, we will discuss the role of multi-modal large models in some application scenarios (taking GPT-4 as an example).

\subsection{Works Based on Specific Generative Networks}

Although there are currently many works on the applications of large multi-modal models in many scenarios, there is still little benchmark based on large multi-modal models represented by GPT-4 used for artificial intelligence-generated content evaluation. Current works mainly focus primarily on specialized generative networks. 

\subsection{Generative Application}

\textbf{Painting Design} 1) Creative Inspiration and Conceptualization: Artists can use text-to-image models like GPT-4 to generate visual ideas based on descriptive text~\cite{brade2023promptify}. 2) Rapid Prototyping of Ideas: For painters and designers, quickly visualizing a concept or scene from a text description can significantly speed up the creative process~\cite{ko2023large}. 3) Assisting in Complex Compositions: For intricate designs or compositions, GPT-4's ability to generate images from text can help artists visualize complex scenes or elements that might be challenging to conceptualize otherwise~\cite{bubeck2023sparks}. 4) Diversifying Artistic Styles and Techniques: GPT-4 can offer a range of artistic styles and techniques based on the description, enabling artists to explore different aesthetics beyond their habitual approach.

\textbf{Medical Imaging and Visualization} Medical imaging and visualization are essential components in diagnosing and treating diseases, playing a pivotal role in modern healthcare~\cite{mazurowski2023segment,zhang2023medical,chakraborty2023overview,jiang2023review,liu2022multi,rueckert2019model,wang2023media}. Recently, there has been a growing interest in leveraging Generative Pre-trained Transformer (GPT) models, particularly in medical imaging and visualization. These advanced AI models are being explored for their potential to transform and enhance medical diagnostics and education. However, in the specialized field of medical imaging, the accuracy and authenticity of an image's texture are critical~\cite{zhang2023minimalgan}. The texture, which includes the fine details and patterns seen in medical images, is crucial for accurate diagnosis and understanding of various medical conditions. While GPT-4 has made significant strides in text-to-image generation, its current capabilities still need to improve in producing the high level of detail and accuracy required in medical images.

\textbf{Fashion Design and Visualization} For fashion designers, it is essential to create highly detailed and textured visual representations of garments, accessories, and conceptual fashion designs. While GPT-4 represents a significant leap forward in text-to-image generation, current capabilities still need to be revised to achieve the exacting level of detail and precision often required in the fashion industry.

\section{Method}

\subsection{Original Data}

\begin{figure}[h]%
\centering
\includegraphics[width=1\textwidth]{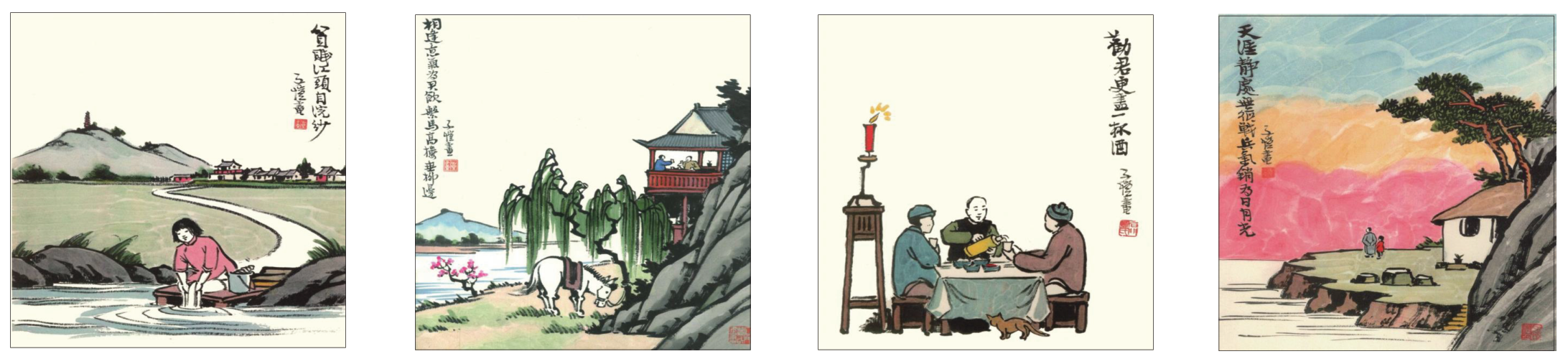}
\caption{These images represent a curated selection from the original dataset, specifically chosen for their ability to accurately convey the appearance and intricate details inherent in the data. Each image serves as a visual representation, offering a clear and detailed glimpse into the characteristics.}\label{selected_example}
\end{figure}

For an in-depth evaluation of GPT-4 and texture feature fidelity in image synthesis, we selected 301 high-quality poem-painting from an influential modern Chinese artist, Feng Zikai, as the original data~\cite{li2021paint4poem}. The data is collected from two books~\cite{zikai1,zikai2} that contain Feng Zikai’s paintings. The illustrations featured in these books are vividly colored and exemplify high quality, making them prime examples of Feng Zikai's artistic style. Accompanying each painting is a poem that serves as its inspiration, accompanied by a modern Chinese interpretation and additional texts that provide context for a better understanding of the poem. This collection comprises 301 unique poem-painting pairings. The dataset for each pairing includes various metadata elements: PoemID, PoemText, PoemTitle, PoemDynasty, PoemAuthor, Explanation, Commentary, and PaintingID. These metadata elements denote the poem's unique identifier, its text, title, the dynasty during which it was composed, the author's name, annotations on complex words, an analysis of the poem, and the identifier for the associated painting. Figure~\ref{selected_example} displays some illustrative examples of this dataset.

It is very appropriate and beneficial to use such kind of data to study the differences between ai-generated and human-created images, specifically for the following principal reasons: 1) Variety of Styles and Techniques: These artworks enable a thorough analysis of GPT-4’s image-generating capabilities because they represent true human artistic expression. 2) Quality of Artistry: The high level of craftsmanship and artistic expression in these works offers a high standard for comparison. It challenges GPT-4 to meet or mimic the complexity and subtlety found in these masterpieces. 3) Consistent Reference Point: Using unified, well-known artworks ensures a consistent reference point for study. Since these works are easily recognizable and their attributes are well-documented, they provide a stable basis for comparison. 4) Open Access: These artworks are available through open-access initiatives. This makes them legally and easily accessible for research purposes without the complications of copyright restrictions that newer or less-known artworks might have.

\begin{figure}[h]%
\centering
\includegraphics[width=1\textwidth]{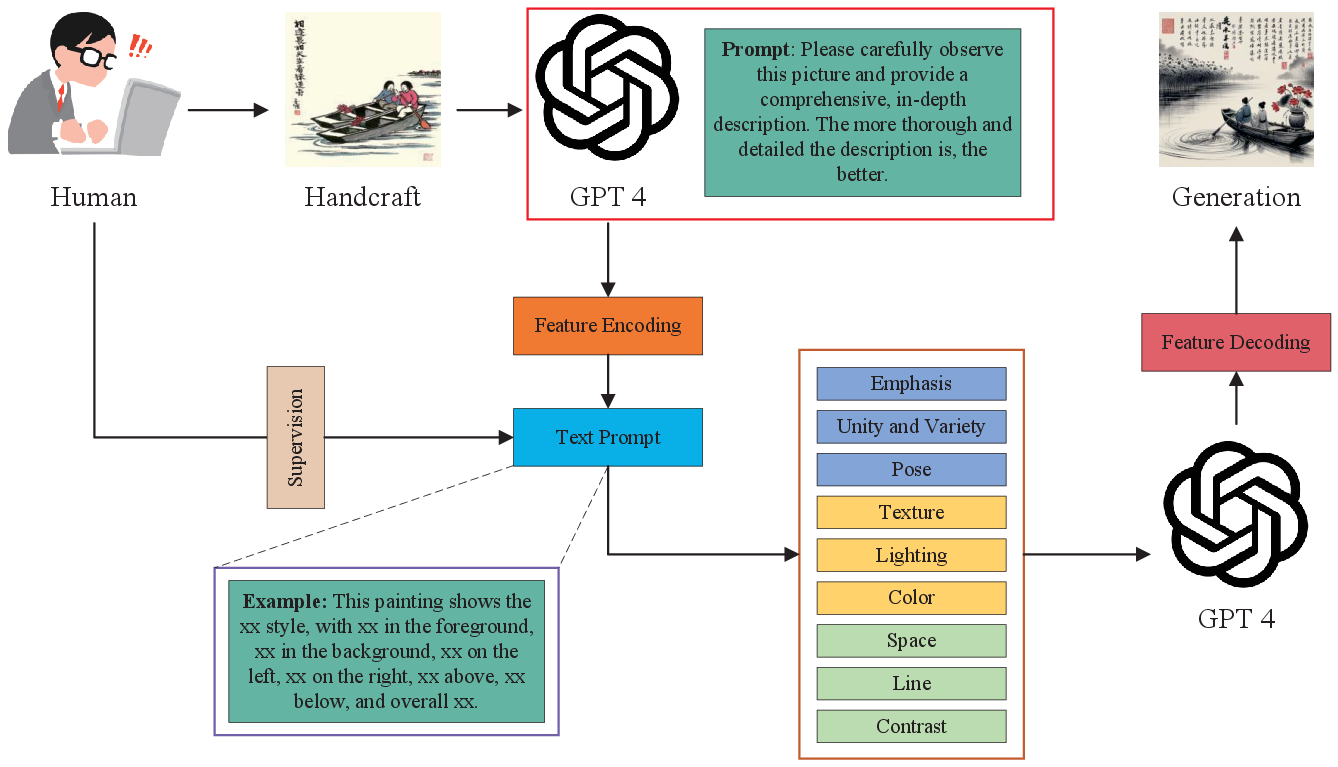}
\caption{This is the overview of the proposed method.}\label{fig2}
\end{figure}

\subsection{Benchmark Creation}

In order to define our benchmark precisely and provide a more detailed and nuanced understanding and analysis, we define an "encoding-decoding" method with GPT-4, focusing on generating deepfake images that closely mimic the attributes of the source images. This method is divided into two distinct phases:

\textbf{Encoding Phase} In this initial step, we input the source images into GPT-4. The GPT-4 then analyzes these images, focusing on identifying and understanding all their features. This analysis culminates in the creation of detailed feature description prompts. These prompts encapsulate the critical elements of the source images, such as color schemes, textures, shapes, and any specific details that stand out. After obtaining the corresponding descriptions, we also performed certain post-processing, manually checking and calibrating the generated descriptions to ensure that these descriptions have sufficient details and applicability while respecting the original outputs of GPT-4.

\textbf{Decoding Phase} The second phase involves reintroducing the previously generated description prompts into GPT-4. This step is crucial as it directs the GPT-4 to leverage these detailed prompts to recreate the target picture's characteristics. The objective here is not just to produce a visual similarity but to recurrence the essence of the original image as closely as possible. This process tests GPT-4's ability to utilize complex descriptive data to generate images highly similar to the source, both in appearance and subtler, nuanced ways.

By employing this two-step encoding and decoding approach, we aim to evaluate the boundaries of GPT-4's capabilities in image processing and recurrence and a level of precision and similarity that blurs the line between the original and the AI-generated image. This method offers a comprehensive framework for analyzing the effectiveness and accuracy of GPT-4 in creating deeply simulated images.

\section{Experiment}

\begin{figure}[h]%
\centering
\includegraphics[width=1\textwidth]{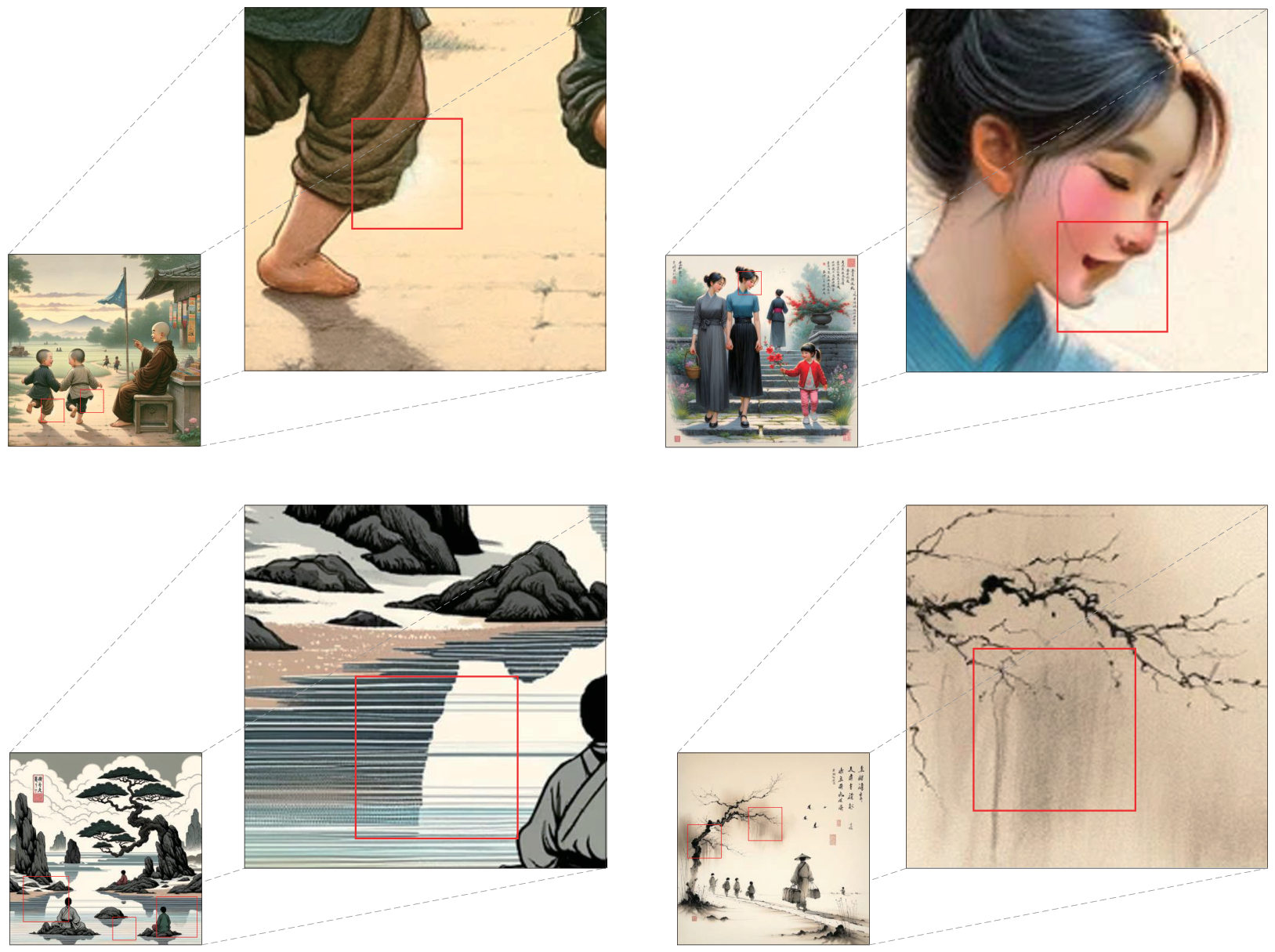}
\caption{Some common textural characteristic errors in images lead to distinct visual anomalies. These anomalies often manifest as irregular white spaces, color confusion, noise streaks, and artifacts, distorting the image's overall texture and composition.}\label{experiment1}
\end{figure}

\begin{figure}[ht]%
\centering
\includegraphics[width=1\textwidth]{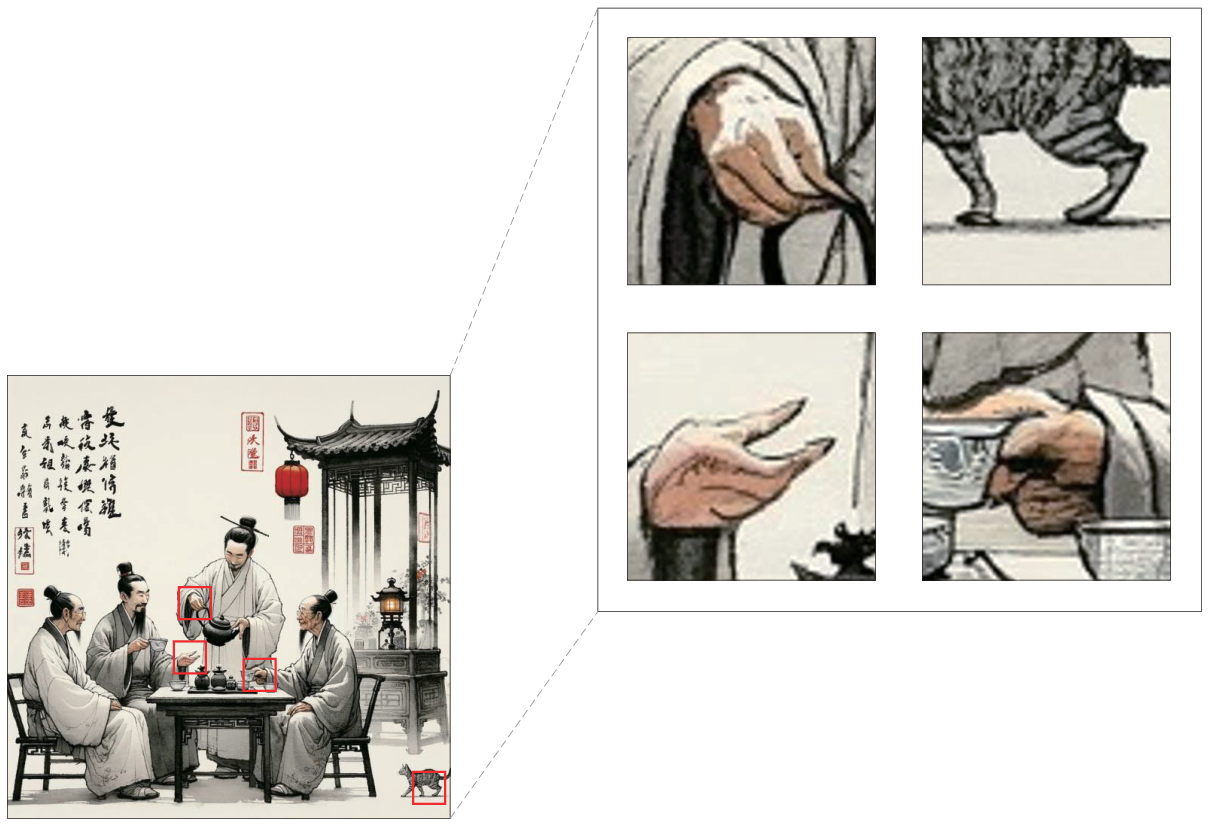}
\caption{Take the selected image as an example. The generated images generally need better processing of details, blur, noise in detail points, and some factual errors that obviously will not occur in humans. The number of human fingers and animal limbs can represent the most obvious ones. , which also proves to a certain extent that the machine intelligence of GPT-4 has flaws in judging the objective facts of nature and lacks prior knowledge of the physiological anatomy of objects.}\label{experiment2}
\end{figure}

\subsection{Qualitative analysis}

To assess the proposed benchmark, we implement the qualitative analysis. The qualitative analysis highlights the critical points of the generated images' features, which could be minor or significant differences from the human-crafted pictures. These points are the key components to justify whether a picture is generated by GPT-4 or human.

\textbf{Unusual Patterns or Textures} Based on established benchmarks, most images produced by GPT-4 showcase distinctive features that set them apart from human-created artwork. These characteristics include patterns or textures that are atypical in human-drawn images. For instance, many GPT-4-generated images display content or patterns not usually observed in human-created images. Additionally, the texturing in many images is unconventional, showing points that differ from common textures. Moreover, there are areas within the image where the level of detail or the style diverges noticeably from the rest of the work. This inconsistency manifests as patches where the image appears either more blurred or sharply defined than its surroundings or where the stylistic approach changes abruptly, breaking the visual continuity seen.

\textbf{Check for Anomalies in Anatomy or Perspective} GPT-4 systems occasionally encounter difficulties when dealing with intricate concepts such as human anatomy and perspective, including limbs that are positioned unnaturally, facial features that appear disproportionate, or scenes that demonstrate incorrect perspective, leading to a lack of realism or visual coherence. Such errors disrupt the overall aesthetic and accuracy of the image.

\textbf{Consistency and Detail Level} Human artists, with their unique techniques and perspectives, tend to display noticeable variations in the detailing of their artworks, often focusing more intensely on some regions of an image. This individualistic approach results in a distinctive, uneven distribution of detail that reflects the artist's particular style or priorities. In contrast, GPT-4-generated images could be characterized by a remarkably uniform level of detail throughout the piece. This uniformity is devoid of the idiosyncratic nuances typical of human artistry. The GPT-4's approach, driven by algorithms, lacks more subjective emphasis and selective detailing that human artists naturally bring to their works.

\textbf{Artistic Errors vs. Digital Artifacts} Despite their precision and consistency, GPT-4-generated images are prone to specific digital artifacts. These manifest as pixelation, where the image resolution is low, leading to visible pixels, or as unusual blending of elements, where the merging of different components in the image appears unnatural or disjointed.

\textbf{Contextual Understanding} Human artists excel in their nuanced grasp of context and the intricate ways different elements within an image interact and relate to each other. This deep understanding allows them to craft images that are not only visually appealing but also rich in meaning and coherence. They can subtly manipulate color, composition, and symbolism to convey complex themes and emotions, creating a harmonious and thought-provoking whole. GPT-4, while advanced in many respects, still face challenges in comprehending the full depth of these relationships. As a result, it can sometimes produce images where the elements, though individually accurate, need to integrate to form a cohesive narrative seamlessly. These GPT-4-generated images display incongruities in theme, style, or logic, detracting from the overall sense of harmony and purpose that characterizes human-created artwork. This gap highlights the current limitations of GPT-4 in fully replicating the nuanced artistic sensibility of human creators.

\subsection{Quantitative analysis}

The benchmark divides images into two main categories: ”Hand-drawn” (which refers to images created by human) and ”Generated images” (which refers to images generated using GPT-4). We define this challenge as a classification problem, and the specific implementation includes the following steps:

\textbf{Baseline Setting} Deep learning is good at analyzing various features, including texture, style, and other visual elements that might indicate the image’s origin. This approach allows for a more nuanced and practical evaluation of GPT-4’s image generation capabilities, providing a deeper understanding of the current limitations of GPT-4 in reproduct human artistic skills. However, we cannot assume that most problems can be solved directly by applying neural networks, because although data-driven neural networks (deep learning) have excellent classification capabilities, their shortcomings are still obvious. That is: when the amount of available data is insufficient, the model may not be able to fully learn, and may even learn some deviations, allowing the model to learn how to deceive (that is, it can be compared to what people often say about taking shortcuts and taking back doors).

\begin{figure}[ht]%
\centering
\includegraphics[width=1\textwidth]{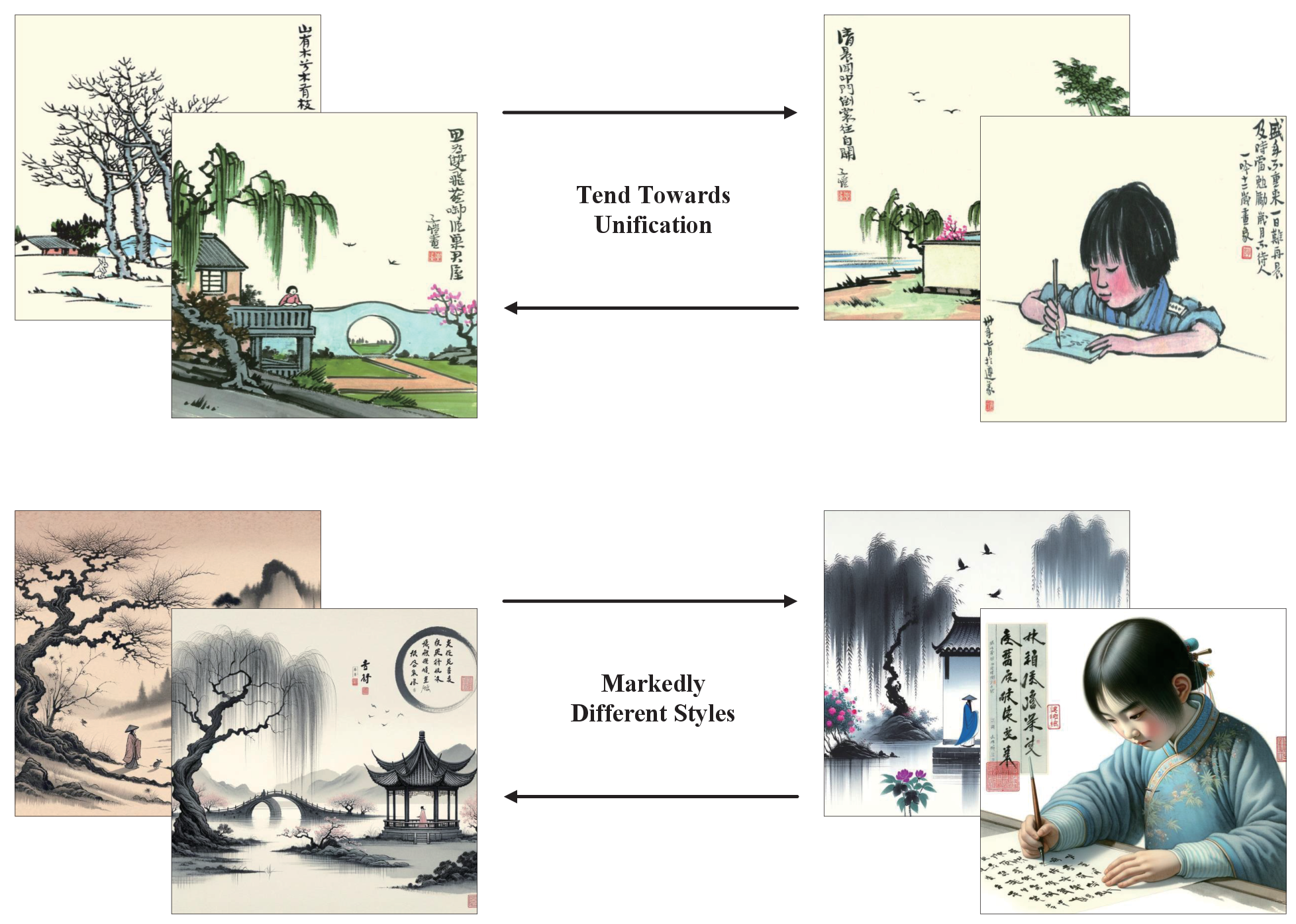}
\caption{Visualization of selected images. It can be seen that there are similarities or identities in styles of human paintings. However, due to the lack of strict constraints, the styles of paintings generated by GPT-4 can become very different, making it extremely easy for neural networks to learn—deviation, thus using the painting style as a basis for judgment.}\label{challenge2}
\end{figure}

The benchmarks we propose can be considered as small samples precisely, i.e. the amount of data may be insufficient. What's more? A more serious problem arises, the real-world images (the paintings) share a similar or same style (The original data was created by the same artist.), as shown in Figure~\ref{challenge2}, but generated images mostly don't strictly adhere to this principle due to the nature of GPT-4. If this task is treated as a typical neural network classification task, the network might learn the style of the real-world sample group and use the trick point: "style" to distinguish between human-created and GPT-4-generated images. This may lead to the model taking shortcuts (ie: cheating) through such tricks, and instead the model simply does not learn the feature differences between the two sets of samples. In other words, the same results can be obtained even with insufficient learning.

Therefore, in order to deal with this problem, we proposed special training and testing strategies, the specific configuration is as follows: 1) Since the focus of this paper is not to design an specific algorithm to supervise the deception behavior of neural networks, the starting point of the strategy is that the simpler and more practical the better. 2) We use diverse backbone networks for classification, which have different parameter amounts and calculation methods, to conduct a comprehensive evaluation of the benchmarks. 3) In order to solve the possible problem of neural network deception, we recommend stopping the training when the network converge and adjusting the partition ratio of the data set to the maximum extent, that is, reducing the proportion of the trainset as much as possible and increasing the proportion of the test set to fully It reflects the learning ability of the network. It can also be observed whether the model makes classification judgments based on learned features or makes judgments by learning shortcuts.

Based on the above strategy as the starting point, we ended up with the following configuration: Training set: validation set: test set divide into 1:1:8, epoch is set to 20, batch size is set to 8, the image resolution was resized to 224$\times$224, using the sgd optimizer and momentum settings is 0.9, weight decay is set to 5e-4, lr decay type is set to cos, and the loss function uses the basic cross-entropy loss function.

\begin{table}[h]
\caption{Experimental Results}\label{compare_results}%
\begin{tabular}{@{}llll@{}}
\toprule
Backbone Network & Human-created Acc (\%) & GPT-4-created Acc (\%) & Average Acc (\%)\\
\midrule
ResNet-18    & 94.17   & 98.33  & 96.25  \\
ResNet-34    & 98.75   & 97.08  & 97.92  \\
ResNet-50    & 97.50   & 99.17  & 98.34  \\
ResNet-101   & 92.50   & 99.58  & 96.04  \\
ResNet-152   & 97.08   & 99.17  & 98.13  \\
\midrule
VGG-11    & 98.33   & 96.67  & 97.50  \\
VGG-13    & 97.50   & 94.58  & 96.04  \\
VGG-16    & 98.33   & 98.33  & 98.33  \\
\botrule
\end{tabular}
\end{table}

\subsection{Evaluation and Analysis}

We directly count the results of batch prediction on the test set. The evaluation of the results directly calculates the ratio of the correct number of predictions to the total number, and uses this result to obtain the score for the benchmark. Then we predict the test set of artificial pictures and the test set of generated pictures respectively to obtain two results, and then calculate the average accuracy of the two to obtain the third result. The Table~\ref{compare_results} shows that most backbone networks are still susceptible to shortcut learning in our tests. With this, we propose the potential challenges of our defined benchmark. Moreover, we would like to provide some insights and ideas.

\textbf{Shortcut Learning} Shortcut learning refers to the tendency of neural networks to select the simplest patterns to complete tasks while ignoring more complex but potentially more critical features. Addressing this problem requires measures to encourage networks to learn deeper and more comprehensive features. Therefore, we mainly put forward two opinions: 1) Certain preprocessing can be carried out on this benchmark so that it cannot take shortcuts completely relying on the clue of color style. For example, a unified style can be processed into a non-uniform color style and, at the same time, Keep other characteristics unchanged. 2) Adjust the task or loss function so that high performance cannot be easily achieved by relying on simple features. For example, we could increase the difficulty of a task by using more fine-grained labels.

In addition to shortcut learning, due to the limitations of open access data, labor costs, and other factors, the size of the current data set is still smaller compared to general vision tasks. Therefore, few-shot learning or zero-shot learning needs to be considered.

\textbf{Few-Shot Learning} Few-shot learning enables models to learn and generalize on tiny amounts of data. This contrasts traditional machine learning methods, which often require large amounts of data to train models to perform well. The critical challenge of few-shot learning is how to effectively extract information from a minimal amount of data and apply this information to new, unseen tasks or samples. Therefore, we propose that particular technologies, such as meta-learning, transfer learning, data augmentation, and model regularization, can be used to explore the low-sample scenario of GPT-4 generated images and artificial image detection.

\textbf{Zero-Shot Learning} Zero-shot learning enables models to recognize, understand, or process data not encountered during their training phase, differing from traditional machine learning methods that typically require training data to encompass all potential scenarios. The fundamental concept of zero-shot learning involves leveraging existing knowledge to infer new, unseen categories or situations. For instance, a model trained for animal recognition might use standard features like limbs and fur to identify an unfamiliar animal it did not encounter during training. This approach is particularly beneficial in scenarios where data is limited. Building on this concept, we propose utilizing knowledge acquired from large-scale synthetic datasets created by advanced generation models, such as Stable Diffusion, to compare samples generated by GPT-4 with human created. However, zero-shot recognition across different domains poses a significant challenge, primarily due to the need to bridge the gap between the domains and the distinct paradigm created by GPT-4.

\section{Discussion and Conclusion}

In this paper, we present a study on the capabilities of current multi-modal large models, mainly focusing on their ability to generate images from text prompts. Our investigation centers on the multi-modal GPT-4 model, examining its ability in image synthesis. Key focus areas include the fidelity of texture features and the discernible differences between generated images. We conduct quantitative and qualitative experiments to assess GPT-4's limitations in image generation. This is analyzed from the perspectives of human vision and specific evaluation metrics, thereby contributing new insights to AI-generated content (AIGC) technology. A pivotal aspect of our research involves creating and analyzing a dataset that pairs hand-crafted artwork with corresponding GPT-4 generated images. This benchmark aims to benchmark and advance the research on the fidelity of AIGC synthesis technology. Through this study, we aim to deepen the understanding of GPT-4's image generation capabilities and its implications for the future of AIGC technology. In our future works, we aim to extend the benchmark further through our dedicated efforts and with the support of other contributors on the open-source platform. This will include but not be limited to expanding the benchmark's scale and refining its formulation. Additionally, we plan to investigate and establish quantitative evaluation standards. These standards will facilitate a more thorough and detailed assessment of the differences between the original images in the DeepArt benchmark and those generated by AIGC.

\bmhead{Acknowledgments}

Thanks to all the individuals, institutions, and groups whose invaluable support was instrumental in completing this research.

\section*{Declarations}

\subsection*{Data Availability}
The benchmark proposed in the paper will be publicly available after being accepted: \url{https://github.com/rickwang28574/DeepArt}

\subsection*{Ethics Declarations}

This research does not involve ethical issues.

\subsection*{Conflict of Interest}
The authors declared that there are no potential conflicts of interest concerning the research, authorship, and publication of this article.

\subsection*{Funding Declaration}

This research was supported by private research funds.

\bibliography{sn-bibliography}

\end{document}